%% file: acl_latex.tex
\documentclass[11pt,breaklinks,hidelinks]{article}

\usepackage[preprint]{acl}

\usepackage{times}
\usepackage{latexsym}
\usepackage{subcaption}  % 加载subcaption包
\usepackage{graphicx}     % 加载graphicx包用于缩放
\usepackage{amsmath,amsfonts}
\usepackage{soul}
\usepackage{url}
\usepackage{paralist} 

\usepackage{enumitem}
\usepackage{graphicx}
\usepackage{caption}
\usepackage{multirow}
\usepackage{amsmath}
\usepackage{amsthm}
\usepackage{booktabs}
\usepackage{algorithm}
\usepackage{algorithmic}
\usepackage[switch]{lineno}
\usepackage{multirow}
\usepackage[T1]{fontenc}
\usepackage[utf8]{inputenc}

\usepackage{microtype}

\usepackage{inconsolata}

\usepackage{graphicx}
\newcommand{\G}{\mathbin{\mathbf{G}}}
\newcommand{\F}{\mathbin{\mathbf{F}}}
\newcommand{\Until}[1]{\mathbin{\mathbf{U}_{#1}}}

\newcommand{\oomit}[1]{}

{\bfseries}{\rmfamily} 
{\bfseries}{\rmfamily} 
{\bfseries}{\rmfamily}
{\bfseries}{\rmfamily}
\newtheorem{mydefinition}{Definition}{\bfseries}{\rmfamily}
{\bfseries}{\rmfamily}

\newcommand{\myparagraph}[1]{\medskip\noindent{\bf #1.}}
\allowdisplaybreaks
\title{Enhancing Transformation from Natural Language to Signal Temporal Logic Using LLMs with Diverse External Knowledge}

\author{
    Yue Fang\textsuperscript{\rm 1,2},
    Zhi Jin\textsuperscript{\rm 1,2}\thanks{Corresponding author},
    Jie An\textsuperscript{\rm 3}\footnotemark[1],
    {\bf Hongshen Chen\textsuperscript{\rm 4}},
    {\bf Xiaohong Chen\textsuperscript{\rm 5}},
    {\bf Naijun Zhan\textsuperscript{\rm 1,2}} \\
    \textsuperscript{\rm 1}School of Computer Science, Peking University, Beijing, China \\
    \textsuperscript{\rm 2}Key Laboratory of High Confidence Software Technologies (PKU), MOE, China \\
    \textsuperscript{\rm 3}National Key Laboratory of Space Integrated Information System, \\Institute of Software, Chinese Academy of Sciences, Beijing, China \\
    \textsuperscript{\rm 4}JD.com, Beijing, China \\
    \textsuperscript{\rm 5}East China Normal University, Shanghai, China \\
    \texttt{y.fang@stu.pku.edu.cn, zhijin@pku.edu.cn, anjie@iscas.ac.cn}
}

\begin{document}
\maketitle
\begin{abstract}
Temporal Logic (TL), especially Signal Temporal Logic (STL), enables precise formal specification, making it widely used in cyber-physical systems such as autonomous driving and robotics. Automatically transforming NL into STL is an attractive approach to overcome the limitations of manual transformation, which is time-consuming and error-prone. However, due to the lack of datasets, automatic transformation currently faces significant challenges and has not been fully explored. In this paper, we propose a NL-STL dataset named STL-Diversity-Enhanced (STL-DivEn), comprising 16,000 samples enriched with diverse patterns. To develop the dataset, we first manually create a small-scale seed set of NL-STL pairs. Next, representative examples are identified through clustering and used to guide large language models (LLMs) in generating additional NL-STL pairs. Finally, diversity and accuracy are ensured through rigorous rule-based filters and human validation. Furthermore, we introduce the Knowledge-Guided STL Transformation (KGST) framework, a novel approach for transforming natural language into STL, involving a generate-then-refine process based on external knowledge. Statistical analysis shows that the STL-DivEn dataset exhibits more diversity than the existing NL-STL dataset. Moreover, both metric-based and human evaluations indicate that our KGST approach outperforms baseline models in transformation accuracy on STL-DivEn and DeepSTL datasets.
\footnote{Project can be found at \url{https://github.com/YueFang0618/STL-DivEn}}
\end{abstract}

\input{1Introduction}

\input{2RelatedWork}
\input{3Preliminary}

\input{4Method}

\input{5Experiment}

\input{6Conclusion}

\section*{Limitations}

Our dataset is currently built using GPT-4 rather than directly derive from requirement documents of
real-world cyber-physical systems. 
Although we have already guided GPT-4 to generate diverse NL-STL pairs, it may still not fully cover the temporal property patterns of real-world cyber-physical systems, or the dataset may be biased. 
This may limit the effectiveness and accuracy of our model when applied to real-world cyber-physical systems.

To address this issue, at least the following approaches can be considered in the future. 
% To address this issue, the following future approaches can be considered:
First, we can extract temporal property patterns from existing real-world cyber-physical systems. 
Second, for specific domains like autonomous driving, we can extract the necessary data from domain-related requirements documentation, e.g., international standards related to AUTOSAR for electronic vehicles.
Furthermore, we can infer possible timing properties and other temporal characteristics of cyber-physical systems by simulating their real interactive environments.
% Furthermore, we can infer possible timing properties and other temporal characteristics of cyber-physical systems by simulation.
In this way, our dataset can be continuously enriched by incorporating human validation to train better models.

\section*{Acknowledgement}
We sincerely thank the anonymous reviewers for their valuable comments and suggestions. This work is supported by the National Natural Science Foundation of China under Grant Nos. 62192731, 62192732, and 62192730.

% Our dataset is currently built by using GPT-4 rather than directly from requirement documents of
% real-world AI-embedded systems. Although we have already guided GPT-4 to generate diverse NL-STL pairs, it may
% still not have fully covered temporal and continuous property patterns of real-world AI-embedded systems, or the
% dataset may be biased. That may limit the effectiveness and accuracy of our model when applied to
% real-world AI-embedded systems.

% To address this issue, there may be at least the following ways in the future. The first is to mine
% temporal and continuous property patterns from the existing real-world AI-embedded systems as many as possible.
% Second, in particular, for data collection in specific domains like autonomous driving, we can extract
% necessary data from domain-related requirements documentation, e.g., the international standard
% of Autosar electronic vehicles. Furthermore, we can also infer the possible patterns of temporal
% and continuous properties for AI-embedded systems by simulating the real interactive environment of AI-embedded systems.
\bibliography{custom}

\clearpage
\appendix
\input{Appendix}

\end{document}

%% file: 1Introduction.tex
\section{Introduction}

% 介绍STL的重要性，以及对LLMs的影响。
%Signal Temporal Logic (STL)~\cite{maler2004monitoring} is a powerful expression tool that describes requirements in AI-embedded systems. 
%Natural language can be ambiguous when expressing complex behavior logic of systems, while 
Signal Temporal Logic (STL)~\cite{maler2004monitoring} provides a flexible and precise framework for specifying requirements in safety-critical cyber-physical systems.
Extending Temporal Logic (TL)~\cite{pnueli1977temporal} by introducing real-time and real-valued constraints, STL can describe not only discrete temporal events but also continuous-time and real-valued dynamic changes.
Therefore, STL, as a powerful expression tool for system design, offers valuable guidance in cyber-physical systems, such as autonomous driving~\cite{maierhofer2020formalization} and robot control~\cite{tellex2020robots}.
% In this way, STL allows users to interact with machine systems more rigorously, transforming ambiguous or complex requirements into precise actions as illustrated in the following example:
But one of the main challenges in leveraging the STL specification is the need to accurately transform the potentially ambiguous and complex constraints expressed in natural language into precise STL logical expressions, as shown in the following example:
\begin{itemize}[leftmargin=0.5cm, nosep] 
\item \textbf{Natural Language:} \\ Whenever the robot detects an obstacle within 1 meter in the first 60 seconds, it should move away from the obstacle and remain at least 1.5 meters away for at least 30 consecutive seconds within the next 50 seconds.
\end{itemize}
\begin{itemize}[leftmargin=0.5cm, nosep] 

\item \textbf{Signal Temporal Logic (STL):} \\ 
\begin{small}$\G_{[0,60]}((d_{\text{obs}}<1)\rightarrow\F_{[0,50]}\G_{[0,30]}(d_{\text{obs}}\geq1.5))$
\end{small}
\end{itemize}
% \begin{itemize}
% \item \textbf{Natural Language:} \\ Whenever the robot detects an obstacle within 1 meter in the first 60 seconds, it should move away from the obstacle and remain at least 1 meter away for at least 30 consecutive seconds within the next 80 seconds.
% \item \textbf{Signal Temporal Logic(STL):} \\ $\F_{[0,60]}(d_{\text{obs}} < 1) \rightarrow \F_{[0,80]} \G_{[0,30]}(d_{\text{obs}} \geq 1)$
% \end{itemize}
% As large language models (LLMs) continue to advance, STL plays a crucial role in supporting LLMs in automatically generating system specifications that meet specific requirements, making artificial intelligence more user-friendly. 
% 最后这一句感觉衔接得有点突兀

% However, it is a huge burden for domain experts to write precise STL formulas directly. 
Writing accurate STL formulas directly is a huge burden for domain experts, as it is both time-consuming and error-prone.
With the development of natural language processing (NLP) technology, researchers have been experimenting with utilizing NLP technology to transform natural language into TL and STL expressions, aiming to improve its automation and accuracy. 
For example,~\citet{LignosRFMK15,ghosh2016arsenal} use predefined pattern formulas to transform natural language sentences into an intermediate representation. Subsequently, by applying a set of predefined rules manually, the intermediate representation is mapped to temporal logic formulas. These approaches require extensive domain expertise and involve a steep learning curve~\cite{kulkarni2013new}. Specifically, they can only be applied to very restrictive structured natural language expressions that match the given patterns. 

In recent years, the remarkable success of deep learning and Large Language Models (LLMs) has spurred growing interest in leveraging them to address the transformation problem from natural language to STL. 
For example, DeepSTL~\cite{HeBNIG22} introduces a grammar-based synthetic data generation technique and trains an attentional translator of English to STL using a transformer-based neural translation technique. NL2TL~\cite{chen2023nl2tl} uses LLMs to help create the Natural Language-Temporal Logic dataset, which is then applied to fine-tune the T5 models. 
However, the proposed transformation methods also face challenges in accurately transforming complex natural language into Signal Temporal Logic. 

To address these challenges, our efforts focus on the following two aspects.
% In recent years, due to the great success of deep learning and LLMs in NLP, more and more attention has been paid to using deep learning and LLMs to solve the translation problem from natural language (NL) to TL. DeepSTL~\cite{HeBNIG22} investigates the transformation from NL to STL by proposing a two-step workflow. The first step is to design a grammar-based generation technique of synthetic data to tackle the challenge of the lack of publicly available informal requirements and formal formulas. The second step is to train an attentional translator of English to STL using a transformer-based neural translation technique. Like DeepSTL, NL2TL~\cite{chen2023nl2tl} also realizes that the lack of datasets is a challenge to the NL-to-TL problem. It proposes to utilize LLMs to assist the dataset creation and then use the set of Natural Language-Temporal Logic (NL-TL) pairs to fine-tune T5 models. In comparison, DialogueSTL~\cite{mohammadinejad2024systematic} is an approach that transforms natural language into STL, ensuring accurate validation through iterative interactions between humans and LLMs.
%First, the open-source NL-STL dataset is not only scarce but also lacks expression diversity,
% and domain-specific terminology, 
%making the development of a high-quality open-source NL-STL dataset highly significant. 
%With the extensive adoption of LLMs, leveraging these models for data synthesis has become a significant research focus. 
Firstly, aiming at developing high-quality and expressively diverse NL-STL datasets to deal with the scarcity of NL-STL datasets, we explore utilizing LLMs to synthesize NL-STL pairs under the guidance of prompts.
However, NL-STL pairs generated by LLMs often closely resemble the examples in the prompts. 
%To address this issue, we introduce the STL-Diversity-Enhanced (STL-DivEn) dataset. 
To ensure diversity and comprehensiveness, we introduce a method for constructing the STL-Diversity-Enhanced (STL-DivEn) dataset. 
We start by handcrafting a seed set of 120 NL-STL pairs, covering both basic and nested logic to serve as the foundation for data augmentation.
% across various domains such as robotics, electronics, and autonomous driving, 
Next, a clustering algorithm is employed to select representative samples from the seed set.
These exemplars are used to guide LLMs in generating new NL-STL pairs, which are then refined using rule-based filters and human validation to ensure diversity and precision.
Finally, the qualified NL-STL pairs expand the seed set and are stored in the STL-DivEn dataset.

Secondly, transformer-based models perform poorly when handling complex natural language transformation tasks. 
Transforming NL sentences into STL formulas remains a challenging task due to the complexity of temporal constraints in the requirements of cyber-physical systems, including nested semantics~\cite{BoufaiedJBBP21}.
%Furthermore, while LLMs excel in text generation tasks, many advanced models, such as GPT-4 and DeepSeek, still face significant limitations in transforming natural language into STL. 
Even many advanced models, such as GPT-4~\cite{achiam2023gpt} and DeepSeek~\cite{liu2024deepseek}, while excelling at text generation tasks, still face limitations in transforming NL into STL.
To address this limitation, we propose a novel transforming framework called Knowledge-Guided STL Transformation (KGST). 
This framework operates in two steps: first, we fine-tune an LLM on NL-STL dataset (e.g., STL-DivEn) and use the finetuned LLM to generate a preliminary STL formula from the natural language input; 
second, the top $K$ similar NL-STL pairs are retrieved from the dataset, and these pairs are referenced as external knowledge; then, GPT-4 is used to evaluate and refine the preliminary STL with the external knowledge to generate the refined STL.
% second, it retrieves the top $K$ similar NL-STL pairs from the dataset, which are referenced as external knowledge to evaluate and refine the preliminary STL using GPT-4 to generate the refined STL.
Experimental results demonstrate that the KGST framework significantly outperforms existing baseline models in both quantitative and human evaluation metrics, showcasing its advantages in STL transformation tasks. 

In general, our contributions are as follows:
\begin{itemize}[nosep]
    \item We develop a dataset, named STL-DivEn, containing 16k high-quality NL-STL pairs using LLMs and manual annotation. Compared to the existing DeepSTL dataset, the statistics show that this dataset exhibits significantly greater diversity.
    % , covering fundamental STL usage and complex nested logic. 
    % It is applicable to multiple domains, including signal processing, robotics, and autonomous driving.
    \item %To address the challenge of transforming natural language to STL,
    % To address the challenges of complex descriptive rules and nested logic in natural language to STL transformation, 
   We propose a Knowledge-Guided STL Transformation (KGST) framework. It substantially improves the accuracy of the NL to STL transformations.
    % by leveraging similar samples in the knowledge base for optimization.
    \item The proposed KGST framework demonstrates superior performance not only on the  STL-DivEn dataset but also on the existing DeepSTL dataset. This highlights its versatility and robustness across different datasets.
\end{itemize}

%% file: 2RelatedWork.tex
\section{Related Work}

% This section contains the work on STL formulas generation, Instruction Tuning and the functions to build instruction tuning dataset.

% \subsection{From Natural Language to TL and STL}
\myparagraph{From Natural Language to TL and STL}
Many researchers have tried to transform natural language sentences into Temporal Logic formulas~\cite{dwyer1999patterns,vzilka2010temporal,ghosh2016arsenal,santos2018formal,CoslerHMST23}. 
% For example, ~\cite{dwyer1999patterns} provided a set of TL-based formulas that capture the typical patterns in the design of concurrent and reactive systems for encoding the property formulas. 
For example,
~\citet{vzilka2010temporal} transform the properties which are specified by controlled English to TL formulas using syntax and grammatical dependency parsing techniques.
~\citet{santos2018formal} also define a controlled natural language to specify how a system model interacts with its environment, and sentences in this controlled language are automatically transformed into TL using predefined rules. 
% ~\cite{ghosh2016arsenal} presented the ARSENAL framework which can be used to translate natural language requirements into formal representations by using NLP techniques like n-gram, dependency parsing, etc and then the latter can be converted into various formalisms, including TL. 
% While these works focus on translating controlled natural language into formal formulas including TL by using mainly pattern-based or rule-based ways which inevitably are domain-specific or can only work for restricted scenarios.
Nl2spec~\cite{CoslerHMST23} derives formal formulas from unstructured natural language using LLMs combined with human corrections.
% However, this approach relies on user feedback and requires manual intervention, which can increase the complexity of usage.
However, these TL-specific approaches cannot be directly applied to STL, as STL involves real-time and real-valued constraints that exceed the expressiveness of TL.

As STL is widely used in academia and industry~\cite{MadsenVSVDWDB18}, several efforts have been made to transform natural languages into STL~\cite{HeBNIG22,chen2023nl2tl,mao2024nl2stl,mohammadinejad2024systematic}.
For instance, DeepSTL~\cite{HeBNIG22} utilizes grammar-based techniques to synthesize data, which is then used to train Transformer models for transformation.
% 合成的数据去训练Transformers模型。
NL2TL~\cite{chen2023nl2tl} fine-tunes T5, trained on lifted Natural Language-Temporal Logic (NL-TL) datasets created by LLMs to perform transformation. 
However, synthetic data generated from specific templates do not capture the full diversity of the real-world language. 
In addition, DialogueSTL~\cite{mohammadinejad2024systematic} transforms natural language task descriptions into accurate STL formulas through user interaction and reinforcement learning, but relies on user feedback, increasing the complexity of usage. To address the insufficient dataset and inefficiencies in transformation, we introduce a new comprehensive dataset and propose a framework to improve the transformation from natural languages to STL. 

%\vspace{-0.1cm}
% \subsection{Instruction Dataset Construction}
\myparagraph{Instruction Dataset Construction}
The generation of instruction datasets involves both manual annotation and synthesis using LLMs.
Manual annotation includes designing prompts and labeling them based on human expertise~\cite{srivastava2023beyond,conover2023hello,zheng2023lmsys,zhaowildchat,zhou2024lima,kopf2024openassistant}. 
However, obtaining high-quality data only through manual annotation can be costly. 
With the growing use of LLMs, research is shifting toward generating data using LLMs, reducing reliance on manual annotation. 
For example,~\citet{taori2023stanford,wang2024codeclm,sun2024principle} start with a small set of seed instructions, which are then expanded using in-context learning to generate diverse instruction-response pairs. 
However, these methods often struggle with ensuring sufficient diversity in the generated data. 
To address this, strategies such as iterative generate-filter pipelines~\cite{wang2023self} and cluster-based data selection~\cite{koksal2024longform} have been proposed. Additionally, WizardLM~\cite{xu2023wizardlm} introduces an instruction evolution paradigm to enhance diversity by increasing the complexity of new instructions.
In our work, the STL-DivEn dataset is created using manual annotation to generate a small set of high-quality seeds. 
LLMs are then used with carefully designed instructions to generate various NL-STL pairs, followed by rigorous validation to ensure consistency.

%% file: 3Preliminary.tex
\section{Signal Temporal Logic} \label{sec:preliminary}
STL
is widely adopted as a specification formalism for cyber-physical systems. 
For example, 
%in model-based design of CPS, 
\emph{Automatic transmission (AT)}, a widely-used benchmark~\cite{ARCHCOMP20Falsification,ARCHCOMP21Falsification,ARCHCOMP22Falsification}, is a transmission controller of automotive systems. 
It continuously outputs the \textsf{gear}, \textsf{speed} and \textsf{rpm} of the vehicle. 
One of its safety requirements is as follows: 
\emph{In the following 27 time units, whenever the speed is higher than 50, the rpm should be below 3000 in three time units.} 
STL can represent such real-time and real-valued constraints.

Let $\mathbb{R}$ denote the set of real numbers. $\mathbb{R}_{\geq 0}$ and $\mathbb{R}_+$ represent the nonnegative and positive real numbers, respectively. Let $\mathbb{N}_+$ be the set of positive integer numbers. 

Let $T\in \mathbb{R}_+$ be a positive real number, and let $d\in\mathbb{N}_+$ be a positive integer. A \emph{$d$-dimensional signal} is a function $\mathbf{v} \colon [0,T] \to \mathbb{R}^d$,
where $T$ is called the \emph{time horizon} of $\mathbf{v}$. Given an arbitrary time instant $t\in[0, T]$, $\mathbf{v}(t)$ is a $d$-dimensional real vector; each dimension concerns a signal \emph{variable} that has a certain physical meaning, e.g., \textsf{speed}, \textsf{rpm}, \textsf{acceleration}, etc. In this paper, we fix a set $X$ of variables and, without ambiguity, we call a variable a signal ($1$-dimensional signal). 
%assume that a signal $\mathbf{v}$ is \emph{spatially bounded}, i.e., for all $t\in[0, T]$, $\mathbf{v}(t)\in\Omega$, where $\Omega$ is a $d$-dimensional hyper-rectangle. 

\begin{mydefinition}[STL Syntax]
% In STL, the \emph{atomic propositions} $\alpha$ and the \emph{formulas} $\varphi$ are respectively defined as follows:
In STL, atomic formulas $\alpha$ and formulas $\varphi$ are inductively defined as follows:
\begin{small}
\begin{gather*}
 \alpha \,::\equiv\, f(x_1, \dots, x_K) > 0 \\
 \varphi \,::\equiv\,
\alpha \mid \bot
\mid \neg \varphi 
\mid \varphi_1 \wedge \varphi_2
\mid \G_{I}\varphi
\mid \F_{I}\varphi
\mid \varphi_1 \Until{I} \varphi_2
%\mid \varphi_1 \Since{I} \varphi_2
\end{gather*}
\end{small}
where $f$ is a $K$-ary function $f:\mathbb{R}^K \to \mathbb{R}$, $x_1, \dots, x_K \in X$, and $I$ is a closed non-singular interval in $\mathbb{R}_{\geq 0}$, i.e.,\ $I=[l,u]$, where $l,u \in \mathbb{R}_{\geq 0}$ and $l<u$.
$\G, \F$, and $\Until{}$ are temporal operators, which are known as \emph{always}, \emph{eventually} and \emph{until}, 
%and \emph{Since}
 respectively. The always operator $\G$ and eventually the operator $\F$ are two special cases of the until operator $\Until{}$, which can be defined by $\F_{I}\varphi\equiv\top\Until{I}\varphi$ and $\G_{I}\varphi\equiv\lnot\F_{I}\lnot\varphi$.
Other Boolean connectives such as $\lor, \rightarrow$ are introduced as syntactic sugar, i.e.,  
%$\top \equiv\neg\bot$, 
$\varphi_1\lor\varphi_2\equiv \neg(\neg\varphi_1\land\neg\varphi_2)$, $\varphi_1\to\varphi_2 \equiv \neg\varphi_1 \lor \varphi_2$. 
\end{mydefinition}

The \emph{Boolean semantics} of an STL formula can be described in a satisfaction relation $(\mathbf{v},t) \models \varphi$, which represents the signal $\mathbf{v}$ that satisfies an STL formula $\varphi$ at time $t$:
\begin{small}
\begin{align*}
   & (\mathbf{v},t) \models \alpha  &  \Leftrightarrow \quad & f(\mathbf{v}(t))\geq 0 \\
   & (\mathbf{v},t) \models \neg\varphi & \Leftrightarrow  \quad & (\mathbf{v},t) \not\models \varphi \\
   & (\mathbf{v},t) \models \varphi_1 \wedge \varphi_2 & \Leftrightarrow  \quad & (\mathbf{v},t) \models \varphi_1 \wedge (\mathbf{v},t) \models \varphi_2 \\
   & (\mathbf{v},t) \models \G_{[l,u]} \varphi & \Leftrightarrow  \quad & \forall t'\in [t+l,t+u].\, (\mathbf{v},t') \models \varphi \\
   & (\mathbf{v},t) \models \F_{[l,u]} \varphi & \Leftrightarrow  \quad & \exists t'\in [t+l,t+u].\, (\mathbf{v},t') \models \varphi \\
   & (\mathbf{v},t) \models \varphi_1 \Until{[l,u]} \varphi_2 & \Leftrightarrow  \quad & \exists t'\in [t+l,t+u].\, (\mathbf{v},t') \models \varphi_2 \\ & & & 
 \wedge \forall t'' \in [t, t']. \, (\mathbf{v},t') \models \varphi_1
\end{align*}
\end{small}

Now, we can formally specify the above \emph{AT safety requirement} by the following STL formula: 
\begin{small}
$$\G_{[0,27]}(\textsf{speed} > 50 \to \F_{[1,3]}(\textsf{rpm} < 3000)).$$
\end{small}

Note that \emph{nested} STL formula refers to an STL formula where temporal operators are applied within the scope of other temporal operators. 

% \begin{mydefinition}[Nested STL formula] 
% We call an STL formula $\varphi$ nested if it can be written in one of the following forms:
% \[
% \varphi = F[a,b]\varphi_1
% \]
% \[
% \varphi = G[a,b]\varphi_1
% \]
% \[
% \varphi = \varphi_1 U[a,b] \varphi_2
% \]
% where $\varphi_1$ in the first two cases and at least one of $\varphi_1$ and $\varphi_2$ in the third case include temporal operators. In addition, $\varphi_1$ in the first two cases and $\varphi_1$, $\varphi_2$ in the third case are called the argument(s) of the STL formula $\varphi$.

% Examples of nested STL formulas include:
% \[
% F[a_1,b_1]G[a_2,b_2]\mu
% \]
% \[
% G[a_1,b_1]F[a_2,b_2]\mu
% \]
% \[
% \mu_1 U[a_1,b_1](G[a_2,b_2]\mu_2 \land F[a_3,b_3]\mu_3)
% \]
% and so on.

% \end{mydefinition}
%\textcolor{red}{to be continued.}

%% file: 4Method.tex
\section{Approach} \label{sec:approach}

% 首先，人工构造了一部分初始数据，并使用这些数据指导语言模型生成更多数据。接着，使用UCB分数从这些种子中选择得分最高的3个，并以这些种子为指导，进一步生成新的NL-STL数据。然后，判断新生成的NL-STL与种子库中的NL-STL的ROUGE分数是否低于预设阈值。如果低于阈值，则将其加入种子库并进行人工核查，将核查通过的数据集作为一个新的数据集。

% 将上述数据集划分为训练集和测试集。使用训练集对开源大模型进行微调，模型的输入为自然语言，输出为对应的STL规范。接下来，针对一个新的自然语言输入，首先通过微调后的大模型生成一个初步的STL（Preliminary STL）。然后，使用GPT-4作为优化器（或校正器，可更换名称），从训练集中筛选出与输入自然语言相似度最高的三个NL-STL对，并将它们与当前输入的自然语言和生成的Preliminary STL一起输入GPT-4。GPT-4根据上下文信息对Preliminary STL进行校验和优化，最终输出经过改进的STL规范作为结果。

In this section, we first present our approach for constructing the STL-Diversity-Enhanced (STL-DivEn) dataset, which combines manual annotation and LLMs to generate diverse, high-quality data. 
Second, we introduce the Knowledge-Guided STL Transformation (KGST) framework to further enhance performance in STL transformation.

% 这里的数据集，结合了人工和大模型感觉要调整一下。
\subsection{Dataset Construction}

\begin{figure*}[!t]
    \centering
    \includegraphics[width=1\linewidth]{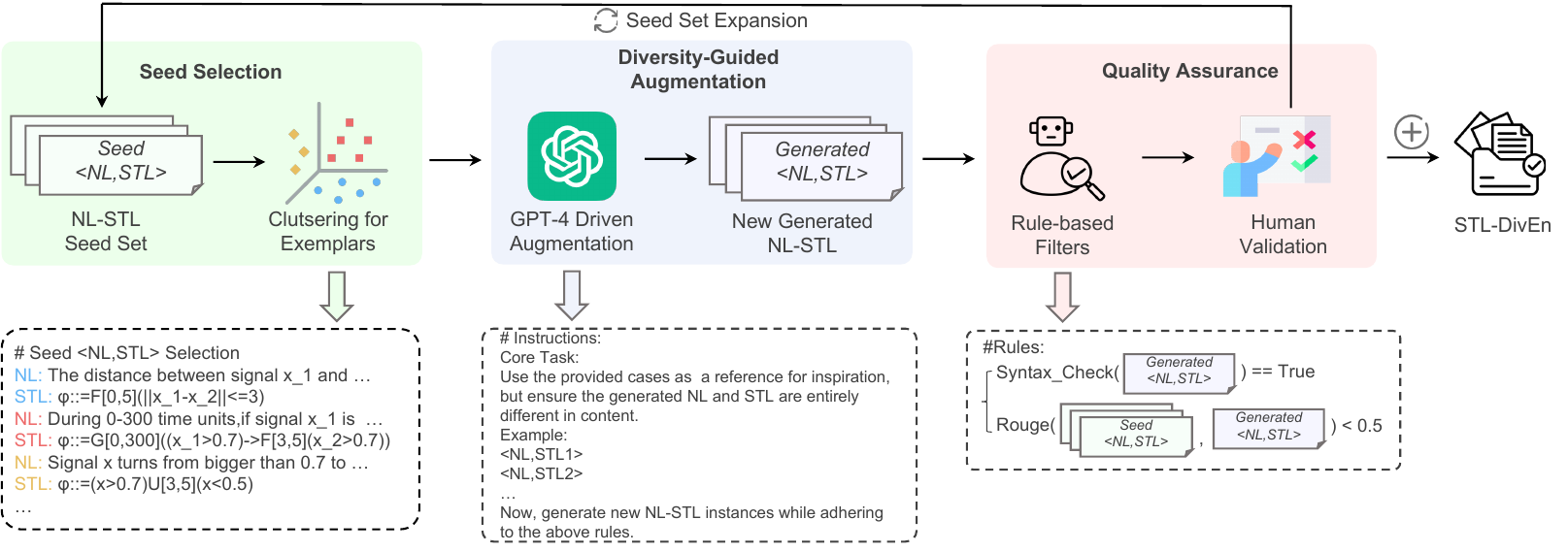}
    %\vspace{-1mm}
    \caption{The pipeline of STL-DivEn construction. We first handcraft a set of seed NL-STL pairs. Next, representative NL-STL pairs are selected by clustering to guide GPT-4 in data augmentation. The newly generated NL-STL pairs pass through rule-based filters and human validation. Finally, verified pairs are added to the STL-DivEn dataset and seed set for the next round generation.}
    %\vspace{-3mm}
    \label{fig:dataset_construction}
\end{figure*}

% 写一下总览，分成哪几步。

% 目前的NL-STL数据集是不足的，并且主要依赖于规则来生成，其中的多样性很低，并且不够全面。在这里，我们提出了一个用于生成NL-STL的数据集的方法。它利用手工制作seed set，聚类选择示例，用指令驱动增强生成，再通过规则过滤和人工验证，最终更新并构建seed set和STL-DivEn数据集。

To build a comprehensive and diverse NL-STL dataset, we follow the steps below: 1) Seed Selection: Manually create an initial set of NL-STL pairs and use clustering algorithm to identify representative seeds, 2) Diversity-Guided Augmentation: Utilize the identified seeds as diverse examples to guide GPT-4 (gpt-4-0125-preview) for augmentation in generating new NL-STL pairs, 3) Quality Assurance: Apply rule-based filters to remove low-quality pairs and human validation to verify semantic consistency, and 4) Dataset Expansion: Add qualified pairs to the seed set and store them in the STL-DivEn database. This pipeline is illustrated in Figure~\ref{fig:dataset_construction}.
% To build a comprehensive and diverse NL-STL dataset, we handcraft a instruction seed set covering basic temporal operators and nested logical operations. Subsequently, we utilize examples from the seed set to guide LLMs in augmentation. To ensure the diversity of these examples, we use K-means clustering to select cluster centers as examples. We then compare the newly generated NL-STL pair with the seed set using ROUGE scores. For pairs that have a Rouge score less than 0.5 with all seeds in the seed set and satisfy the syntax rules mentioned in Section~\ref{sec:preliminary}, we will manually verify the semantic consistency between NL and STL in this pair. Finally, the qualified NL-STL pairs are added to the seed pool to guide the next round of generation and are also stored in our STL-DivEn database.

% 为了构建一个全面且多样化的NL-STL数据集，我们手工编写了一个基础数据集，涵盖了基本的时间运算符和嵌套逻辑运算。随后，我们需要一些例子来指导LLM进行增强。为了确保这些例子更加多样化，我们使用K-means聚类来选择聚类中心作为例子。我们随后用ROUGE得分比较了新生成的NL-STL对与现有数据集的差异。对于那些与现有数据不同且满足第三节提到的语法规则的对，我们会通过人工方式验证STL与NL之间的语义一致性。最后，合格的NL-STL对被添加到种子库中，以指导下一轮的生成，同时也存储到我们的STL-DivEn数据库中。

\noindent\textbf{Seed Selection.} Signal Temporal Logic encompasses a variety of complex applications. Without high-quality seeds, the generated data may lack diversity. 
Therefore, the first step is to build a seed set, which includes natural language descriptions and corresponding STL formulas covering both nested and basic logic, as well as applications in fields such as autonomous driving, robotics, and electronics.
To ensure both comprehensiveness and accuracy, these initial NL-STL pairs are manually created. The seed set is created by 6 domain experts, two from each field, resulting in a total of 120 NL-STL pairs, with 40 pairs from each field.
% involves constructing a basic seed set consisting of pairs of natural language (NL) descriptions and their corresponding Signal Temporal Logic (STL) formulas. 
% These initial NL-STL pairs are handcrafted to ensure comprehensiveness and accuracy. Each pair is carefully crafted by 6 domain experts, covering a wide range of STL applications, including basic temporal operations and more complex nested logic structures, with a total of 120 NL-STL pairs.
% The term nested STL refers to formulas in Signal Temporal Logic (STL) that contain nested temporal operators~\cite{yu2024continuous}.

When using GPT-4 to generate new NL-STL pairs, selecting appropriate examples is crucial as the generated NL-STL pairs tend to mimic the provided examples. To ensure diversity, we employ the k-means~\cite{hartigan1979k} to cluster five centers from the seed set, and then use these centers as examples to guide the GPT-4 in data augmentation.

We use the Sentence-Transformers~\cite{reimers-2019-sentence-bert} to map NL-STL pairs into a high-dimensional vector space, determining the cluster centers. This approach prevents any single category of NL-STL pairs from dominating the generated data.
% 第一步涉及构建一个基础的种子数据集，该数据集由自然语言（NL）描述与其对应的信号时序逻辑（STL）规范配对而成。这些初始的NL-STL对是通过人工生成的，确保了高质量和准确性。每一对均由领域专家精心构建，涵盖了广泛的STL应用，包括基本的时序运算和更复杂的嵌套逻辑结构，共包含120条NL-STL对。
% 所谓嵌套逻辑结构，是指在信号时序逻辑（STL）中，包含嵌套时序算子的公式。

% 在利用大模型生成新的NL-STL对时，由于新生成的NL-STL对会模仿给定的示例，因此示例的选择至关重要。为了确保生成NL-STL的多样性，我们采用了k-Means聚类算法，从数据中计算出5个聚类中心的NL-STL对，并将它们作为示例去指导语言模型。

% 在k-Means聚类过程中，我们使用Sentence-Transformers模型将NL-STL对映射到高维向量空间，从而确定聚类中心。这种方法有效避免了某一类NL-STL对在生成数据中占据过高比例的问题，从而实现了数据的多样性。

\noindent\textbf{Diversity-Guided Augmentation.} 
After selecting the most representative NL-STL pairs, the next step is to generate new NL-STL pairs based on these seeds to expand the dataset. 
The five chosen NL-STL instruction seeds are used as input examples for GPT-4, with evolution prompts guiding GPT-4 to generate new NL-STL pairs. The prompts can be found in the Appendix~\ref{sec:GPT-4_generate}.

\noindent\textbf{Quality Assurance.}
Since GPT-4 may produce incorrect NL-STL pairs, including those with syntax errors, redundancy with the seed set, or inaccurate semantics, we employ rule-based filtering and human validation to ensure the quality of the dataset.

In detail, rule-based filtering is applied in two stages. The first stage applies the syntax check algorithm to eliminate NL-STL pairs that do not adhere to the syntax rules outlined in Section~\ref{sec:preliminary}.
Each NL-STL pair is then compared to the existing data in the seed set by calculating their Rouge scores~\cite{lin2004rouge}. If the Rouge score between a new NL-STL pair and all existing seed pairs is below 0.5, the new pair is considered to exhibit sufficient diversity.

Next, the NL-STL pairs that pass the rule-based filtering undergo human validation to ensure consistency between the natural language and STL specifications. Seven annotators who have been trained in STL usage and expressions spend two months conducting the annotation.
% 在第 2 步中选定了最有价值的 NL-STL 种子后，下一步是基于这些选定的种子生成新的 NL-STL 对，从而扩展数据集。
% 将选定的种子输入到语言模型中，用变异指令指导语言模型生成新的 NL-STL 对，并进行基于规则的筛选，筛选掉语法错误的NL-STL对。在生成新数据后，每个 NL-STL 对都会与种子池中的现有数据计算和它们之间的Rouge分数。
% 如果新数据与现有的所有种子之间的 ROUGE 得分低于预设阈值，则认为这些新生成的数据是有价值的，并将其纳入种子池。这些新的 NL-STL 对随后将作为下一轮种子生成的指导，确保数据集通过迭代不断进化和改进。
% the newly generated data is considered valuable and added to the seed pool. 
% These new NL-STL pairs will subsequently serve as guidance for the next round of seed generation, ensuring the dataset evolves and improves iteratively.

\noindent\textbf{Dataset Expansion.} To continuously enhance data diversity, NL-STL pairs filtered through rule-based filtering and human validation are added to the seed set as candidates for guiding the next generation. These pairs are also incorporated into the STL-DivEn dataset, which is organized in a structured format that links natural language expressions to their corresponding STL formulas.

% To continuously enhance dataset diversity, NL-STL pairs filtered through rule-based processing and human validation are added to the seed set as candidates for guiding the next generation. These pairs are also incorporated into the STL-DivEn dataset, which is organized in a structured format that links natural language expressions to their corresponding STL formulas.

% In order to continuously enhance dataset diversity, NL-STL pairs filtered through rule-based processing and human validation are added to the seed set as candidates for guidance the next generation round. These pairs are also incorporated into the STL-DivEn dataset, which is organized in a structured format that links natural language expressions to their corresponding STL formulas.

% 最终，所有新生成的数据都将经过人工验证。该验证过程旨在确保数据符合高质量标准，并且不存在任何错误或不一致之处。为此，我们邀请了7位经验丰富的标注者参与此任务，并花费了2个月的时间完成了对数据集的全面标注工作。这一过程不仅确保了数据的准确性和一致性，还为模型训练提供了坚实的基础。

\subsection{Applying LLMs to Generate Formulas}

\begin{figure}[!t]
    \centering
    \includegraphics[width=1\linewidth]{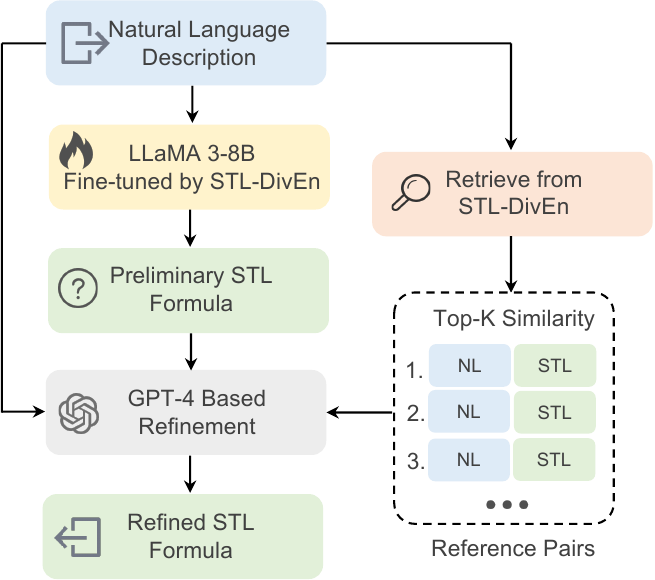}
    %\vspace{-1mm}
    \caption{Architecture of Knowledge Guide STL Transformation (KGST).
    }
    \label{fig:KGST}
    % \vspace{-0cm}
\end{figure}

To enable LLMs to utilize the acquired knowledge more effectively, we structure the NL-STL transformation task as a generate-then-refine process, as shown in Figure~\ref{fig:KGST}.

Specifically, we first fine-tune LLMs such as LLaMA 3-8B on STL-DivEn, enabling them to transform natural language descriptions into preliminary STL formulas.

Next, GPT-4 is employed to refine the preliminary STL formula. Specifically, we select the top $K$ most similar NL-STL pairs from external knowledge (e.g., STL-DivEn) as reference pairs based on the input natural language description using a similarity algorithm, where $K$ is set to 5.
These reference pairs, along with the original natural language description and the preliminary STL formula, are then fed into GPT-4. 

Finally, GPT-4 evaluates and refines the preliminary STL formula based on the reference pairs, generating the refined STL formula.
The prompts used for this process are detailed in Appendix~\ref{sec:KGST_refine}.
% To more effectively convey and utilize the knowledge acquired in LLMs, we structure the NL-STL transformation task in a coarse-to-refine process. 我们首先 as illustrated in Figure~\ref{fig:KGSR}

% Each task instance is defined by three attributes: \textbf{Instruction}, \textbf{Input}, and \textbf{Output}. \textbf{Instruction} consists of detailed instructions for transforming the input natural language text to the corresponding output STL formula. 

% \textbf{Input} is a natural language description that needs to be transformed. This input sequence, along with the \textbf{Instruction}, is fed into the LLMs, enabling the model to generate the required output for the given task. \textbf{Output} is a preliminary STL formula transformed from the input text.

% % 为了更有效地传递和利用LLM中获得的知识，我们将NL-STL翻译任务结构化为序列到序列（seq2seq）的形式，并通过微调LLM来处理该任务。每个任务实例由三个属性定义。Instruction，Input和Output。

% Then, a similarity algorithm is used to select the five most similar NL-STL pairs from external knowledge based on the input natural language description. These selected pairs are then input into GPT-4 along with the input natural language and the generated preliminary STL. GPT-4 validates and refines the preliminary STL based on contextual information, ultimately outputting the refined STL formula as the result.

% % 然后，使用相似度算法根据输入的自然语言从训练集中选择三个最相似的NL-STL对。然后，将这些选定的对与当前输入的自然语言和生成的初步STL一起输入到GPT-4中。GPT-4根据上下文信息验证和优化初步STL，最终输出经过改进的STL规范作为结果。

%% file: 5Experiment.tex
\section{Experiments} \label{sec:experiment}

In this section, we conduct experiments on our proposed dataset and the existing benchmark proposed~\cite{HeBNIG22} to evaluate our methods.

\subsection{Experiment Settings}
We first introduce our empirical settings, including datasets, evaluation measures, baselines and implementation details.

\noindent\textbf{Datasets.} We conduct experiments on two NL-STL datasets, including DeepSTL~\cite{HeBNIG22} and the proposed STL-DivEn Dataset. Specifically, DeepSTL generates STL formulas through randomly sampling from templates and operator distributions, while STL-DivEn is a dataset created using GPT-4 and human annotation. We randomly selected 14,000 samples from each dataset for the training set and 2,000 samples for the test set.
% Among them, DeepSTL introduces an approach that defines a constrained fragment of STL, generating STL formulas through random sampling based on empirically derived templates and operator distributions, and transforming them into English using syntactic strategies. 

% 我们在两个NL-STL数据集上进行实验，包括DeepSTL~\cite{HeBNIG22}和我们提出的STL-DivEn数据集。DeepSTL引入了一种方法，通过基于经验得出的模板和运算符分布进行随机采样，定义STL的约束片段，生成STL公式，并使用句法策略将其翻译成英文。我们从中提取了一个包含16,000个样本的数据集进行实验。STL-DivEn是一个包含16,000个NL-STL对的数据集，基于人工注释和大型语言模型（LLMs）提出。

% DeepSTL generates STL formulas by randomly sampling from a distribution of templates and operators based on empirical statistics, defining restricted STL fragments, and translating them into English using syntactic strategies, resulting in a dataset of 120,000 pairs of STL-English sentences. Our work is based on LLMs and introduces a dataset containing 12,000 pairs of NL-STL data.
% NL2TL constructs a dataset by combining GPT-3 generation with existing data refinement, comprising approximately 28K pairs of enhanced NL-STL data. Our work is based on LLMs and introduces a dataset containing 12,000 pairs of NL-STL data.

\noindent\textbf{Evaluation Measures.} To evaluate the results of STL generation, we utilize both quantitative metrics and human evaluation in our experiment. 
In detail, we use three evaluation metrics: STL Formula Accuracy, Template Accuracy~\cite{HeBNIG22}, and BLEU~\cite{papineni2002bleu}, which are used for the STL generation task. 
STL Formula Accuracy emphasizes strict alignment of symbols and syntax, Template Accuracy evaluates the completeness of logical structures, and BLEU assesses local semantics and phrase-level similarity.
The calculation methods for STL Formula Accuracy and Template Accuracy are provided in Appendix~\ref{sec:Metrics}.
% 为了评估STL生成的结果，我们使用了三种标准评估指标：STL Formula Accuracy，Template Accuracy和BLEU。STL Formula Accuracy注重严格的符号和语法对齐，Template Accuracy关注逻辑结构的完整性，BLEU则偏向于局部语义和短语相似性的评估。

% STL Formula Accuracy: This measures the alignment accuracy between the reference sequence and the predicted sequence at the token level. It is calculated as the ratio of the number of correctly predicted tokens at the same position in the predicted formula compared to the total number of tokens.
% Template Accuracy: This metric first converts both the reference and predicted instances into STL templates and then calculates their alignment accuracy. Compared to formula accuracy, template accuracy is typically higher because when formulas are converted to templates, some potential translation errors (such as incorrect identifiers, constants, or logical operators) may be obscured.
% BLEU (Bilingual Evaluation Understudy): This evaluates the number of 4-grams present in the reference sequence. A perfect score of 1 indicates complete overlap and is used to assess the similarity between the generated translation and the reference translation.

For human evaluation, we randomly selected 100 NL-STL pairs from the test set of STL-DivEn and DeepSTL. Five annotators (all students who have grasped the usage of STL formulas) are required to compare our model with baseline models. They are unaware of which STL formulas are generated by our model and which are generated by the baseline models. 
The annotators evaluate whether the STL formula faithfully reflects the natural language description in four aspects: whether the operators in the STL are correct, whether the values are accurate, whether the generated STL conforms to the syntax rules, and whether the semantics are consistent with the natural language description. The evaluation results are labeled as correct only when all aspects are correct; otherwise, they are marked as incorrect if any aspect is wrong.

% 三位标注者（均为学习过STL公式课程的学生）对我们的模型与基线模型进行比较。他们不知道哪些STL公式是由我们的模型生成的，哪些是来自基线模型的。
% 标注者需要从四个方面检查STL公式是否忠实反映了自然语言描述：STL中的运算符是否正确、值是否准确、语义是否一致，以及生成的STL是否符合语法规则。评估结果被标注为“正确”或“错误”。

% \noindent\textbf{Implementation Details.}
% In this work, we fine-tune our dataset based on LLaMA 3-8B and use GPT-4 to support the refinement. The experiments were conducted on eight 4090 GPUs. We implemented all models using PyTorch, Llama-Factory, and Huggingface's Transformers. 

%在本工作中，我们基于 LLaMA 3-8B 对数据集进行了微调，并使用 GPT-4 作为 Refine 所依赖的大模型。实验在 8 台 4090 GPU 上进行。我们采用 PyTorch、Llama-Factory 和 Huggingface 的 Transformers 实现所有模型。在初始化重写器时，使用 Adam 优化器（Kingma 和 Ba，2017），设置学习率为 5e-6，批量大小为 2，训练 25,000 次迭代。

\noindent\textbf{Baselines and Implementation Details.} We conduct the comparison experiments using five baseline methods: DeepSTL, GPT-3.5\footnote{https://platform.openai.com/docs/models/gpt-3-5-turbo}, GPT-4\footnote{https://platform.openai.com/docs/models/gpt-4-turbo-and-gpt-4}, DeepSeek~\cite{liu2024deepseek}, and Self-Refine~\cite{madaan2024self}. In our experiments, the GPT-4 version is "gpt-4-0125-preview", the GPT-3.5 version is "gpt-3.5-turbo-1106", and the DeepSeek version is "DeepSeek-V3". 
The Self-Refine method involves GPT-4 generating an initial STL formula, followed by refinement using GPT-4's own knowledge. 
DeepSTL uses the Adam optimizer~\cite{kingma2014adam} and is trained with the Transformers model architecture. 
KGST is fine-tuned on LLaMA 3-8B and utilizes GPT-4 for refinement with external knowledge, which is derived from the corresponding training set.
Details on hyperparameter determination are provided in Appendix~\ref{sec:Details}.
% 我们在DeepSTL和STL-DivEn数据集上使用了五个基线模型进行比较，分别是DeepSTL、GPT-3.5、GPT-4、DeepSeek和Self-Refine。在实验中，GPT-4的版本为“gpt-4-0125-preview”，GPT-3.5的版本为“GPT-3.5-turbo-1106”，DeepSeek的版本为DeepSeek-V3。Self-Refine通过GPT-4首先生成初步的STL公式，然后利用GPT-4自身的知识对其进行优化。DeepSTL使用Adam优化算法，根据模型类型（Transformer）设置不同的超参数进行训练。

\subsection{Experimental Results}
In this section, we show our experimental results on the two datasets STL-DivEn and DeepSTL. 

\subsubsection{Metric-Based Evaluation}

The quantitative evaluation results on the STL-DivEn and DeepSTL datasets are shown in Table~\ref{tab:metrics}. 
For the STL-DivEn dataset, our model performs the best (Table~\ref{tab:ours}). 
Across the three metrics, our model achieves scores of 0.5587 for STL Formula Accuracy, 0.5627 for Template Accuracy, and 0.2142 for BLEU, surpassing other models. 
For example, DeepSeek obtains 0.4790, 0.4852, and 0.0791, while GPT-4 obtains 0.4733, 0.4741, and 0.1931 for the respective metrics.

\begin{table}[!t]
    \centering
    \begin{minipage}{\linewidth}  % 设置第一个子表格的宽度
        \centering
        \resizebox{0.9\linewidth}{!}{
\begin{tabular}{lccc}
\toprule
Model & \begin{tabular}[c]{@{}c@{}}STL Formula \\ Accuracy\end{tabular} & \begin{tabular}[c]{@{}c@{}}Template \\ Accuracy\end{tabular} & BLEU \\ \midrule
DeepSTL  & 0.1986 & 0.1883 & 0.0293 \\ 
GPT-3.5 & 0.3018 & 0.3034 & 0.0424 \\
GPT-4 & 0.4733 & 0.4741 & 0.0831 \\
DeepSeek & 0.4790 & 0.4825 & 0.0791 \\
GPT-4+Self-Refine & 0.4422 & 0.4466 & 0.0521 \\
KGST & \textbf{0.5587} & \textbf{0.5627} & \textbf{0.2142} \\ \bottomrule
\end{tabular}
        }
        \subcaption{STL-DivEn}
        \label{tab:ours}
    \end{minipage} 
    %\hfill
    \begin{minipage}{\linewidth}  % 设置第二个子表格的宽度
        \centering
        \resizebox{0.9\linewidth}{!}{
\begin{tabular}{lccc}
\toprule
Model & \begin{tabular}[c]{@{}c@{}}STL Formula \\ Accuracy\end{tabular} & \begin{tabular}[c]{@{}c@{}}Template \\ Accuracy\end{tabular} & BLEU \\ \midrule
DeepSTL & 0.2002 & 0.2916 & 0.3332 \\
GPT-3.5 & 0.2145 & 0.3002 & 0.2249 \\
GPT-4 & 0.2262 & 0.3048 & 0.2881 \\
DeepSeek & 0.2537 & 0.3254 & 0.3982 \\
GPT-4+Self-Refine & 0.2203 & 0.3019 & 0.2682 \\
KGST & \textbf{0.4538} & \textbf{0.4939} & \textbf{0.5686} \\ \bottomrule
\end{tabular}
        }
        \subcaption{DeepSTL}
        \label{tab:deepstl}
    \end{minipage}
    %\vspace{-0.2cm}
    \caption{Metric-based evaluation results.}
    \label{tab:metrics}
    %\vspace{-0.1cm}
\end{table}

For the DeepSTL dataset, as shown in Table~\ref{tab:deepstl}, we also observe that our model achieves the highest scores. It obtains 0.4538 for STL Formula Accuracy, 0.4939 for Template Accuracy, and 0.5686 for BLEU, outperforming all other models. Specifically, DeepSeek obtains 0.2537, 0.3254, and 0.3982, while GPT-4 obtains 0.2262, 0.3048, and 0.2882 for the respective metrics.

Furthermore, we observe a decrease in the performance of the Self-Refine method after refinement. This suggests that refining STL formulas requires external knowledge rather than relying solely on the model's internal capabilities. 
% Additionally, we observed that on both datasets, KGST performs better on the STL-DivEn dataset than on DeepSTL. This indicates that the KGST method has an advantage in handling the complexity and structure of the STL-DivEn dataset. 
In conclusion, our KGST model demonstrates superior performance in generating more accurate STL formulas compared to the baseline models.
% However, on the DeepSTL dataset, our model still demonstrates higher accuracy and generation capability through optimized and enhanced learning strategies, showing that our approach is more robust when dealing with different datasets.

% 在STL-DivEn和DeepSTL数据集上的定量评估结果如表~\ref{tab:metrics}所示。对于STL-DivEn数据集，如表~\ref{tab:ours}所示，我们的模型表现最佳。在三个评估指标上，模型分别取得了STL公式准确率0.5587、模板准确率0.5627和BLEU值0.2142，显著优于其他模型，例如DeepSeek（0.4790, 0.4852, 0.0791）和GPT-4（0.4733, 0.4741, 0.1931）。

% 对于DeepSTL数据集，如表~\ref{tab:deepstl}所示，我们同样观察到我们的模型取得了最高分数。模型在STL公式准确率、模板准确率和BLEU值上分别达到了0.4538、0.4939和0.5686，显著超过了其他模型，例如DeepSeek（0.2537, 0.3254, 0.3982）和GPT-4（0.2262, 0.3048, 0.2882）。

% 此外，我们发现，Self-Refine方法在Refine后，性能相较于GPT-4而言反而出现了下降，这反映了Refine需要借助外部的知识，而不是自身。我们还观察到，在两个数据集上，KGST在STL-DivEn数据集上的表现优于在DeepSTL上的表现，这表明KGST方法在处理STL-DivEn数据集的复杂性和结构方面具有优势。然而，在DeepSTL数据集上，我们的模型通过优化和增强的学习策略，仍然表现出更高的准确性和生成能力，表明我们的方法在处理不同数据集时更加稳健。

\subsubsection{Human Evaluation}

The human evaluation results are shown in Table~\ref{tab:humanevaluation}. 
We use the correctness percentage as a comprehensive evaluation of operator correctness, value accuracy, semantic consistency, and syntax conformity in generated STL formulas.
% Compared with the baseline models, the correctness percentage is used to evaluate the operator correctness, value accuracy, semantic consistency, and syntax conformity of the STL formulas generated by KGST. 
From the results, it can be observed that the evaluators consider the proportion of correct STL formulas generated by our model to be the highest among all methods.
% From the results, it can be observed that the proportion of evaluators who consider our model's generated STL formulas to be correct is the highest. 
For example, on the STL-DivEn dataset, the accuracy of our model is 62.4\%, validating the effectiveness of our KGST model.
% 人工评估结果如表所示。与基准模型相比，从正确百分比来评估KDGR生成STL公式的运算符正确性、值准确性、语义一致性以及语法符合性。从结果中可以看出，认为我们模型生成的STL正确的比例最高。以STL—DivEn数据集上，我们模型的正确率为XX。

\begin{table}[!t]
    \centering % 表格居中
    \resizebox{0.75\linewidth}{!}{
        \begin{tabular}{lcc}
            \toprule
            \multirow{2}{*}{Model} & \multicolumn{2}{c}{Accuracy (\%)} \\ \cmidrule{2 - 3} 
            & STL-DivEn & DeepSTL \\ \midrule
            DeepSTL & 43.4 & 42.0 \\
            GPT-3.5 & 48.4 & 45.6 \\
            GPT-4 & 53.0 & 48.8 \\
            DeepSeek & 55.0 & 49.2 \\
            GPT-4+Self-Refine & 51.2 & 47.0 \\
            KGST & \textbf{62.4} & \textbf{54.6} \\ \bottomrule
        \end{tabular}
    }
    %\vspace{-0.2cm}
    \caption{Human evaluation results.}
    \label{tab:humanevaluation}
    %\vspace{-0.4cm}
\end{table}

\subsection{Analysis}

\subsubsection{Corpus Statistics}

\begin{table}[!t]
    \centering
    \begin{minipage}{\linewidth}
        \centering
        \resizebox{0.95\linewidth}{!}{
            \begin{tabular}{l|cllc|cc|cll}
            \hline
            \multirow{2}{*}{Dataset} & \multicolumn{4}{c|}{\begin{tabular}[c]{@{}c@{}}\#subformula \\ per formula\end{tabular}} & \multicolumn{2}{c|}{\begin{tabular}[c]{@{}c@{}}\#STL oper. \\ per formula\end{tabular}} & \multicolumn{3}{c}{\multirow{2}{*}{\begin{tabular}[c]{@{}c@{}} \#N-gram \\ diversity\end{tabular}}} \\ \cline{2-7}
             & \multicolumn{3}{c|}{avg.} & median & \multicolumn{1}{c|}{avg.} & median & \multicolumn{3}{c}{} \\ \hline
            DeepSTL & \multicolumn{3}{c|}{6.98} & 7 & \multicolumn{1}{c|}{6.98} & 7 & \multicolumn{3}{c}{1.474} \\
            STL-DivEn & \multicolumn{3}{c|}{14.66} & 14 & \multicolumn{1}{c|}{20.04} & 19 & \multicolumn{3}{c}{2.386} \\ \hline
            \end{tabular}
        }
        \subcaption{STL formula statistics: \# subformula for each STL formula, \# operators for each STL formula and \# N-gram diversity of STL formulas.}
        \label{tab:dataset1}
    \end{minipage}
    
    \vspace{0.3cm} % 增加子表格间距

    \begin{minipage}{\linewidth}
        \centering
        \resizebox{0.95\linewidth}{!}{
            \begin{tabular}{l|c|c|cllc|cll}
            \hline
\multirow{2}{*}{Dataset} & \multirow{2}{*}{\#sent.} & \multirow{2}{*}{\#word} & \multicolumn{4}{c|}{\begin{tabular}[c]{@{}c@{}}\#words \\ per sent.\end{tabular}} & \multicolumn{3}{c}{\multirow{2}{*}{\begin{tabular}[c]{@{}c@{}}\#N-gram\\ diversity\end{tabular}}} \\ \cline{4-7}
 &  &  & \multicolumn{3}{c|}{avg.} & median & \multicolumn{3}{c}{} \\ \hline
DeepSTL & 120,000 & 265 & \multicolumn{3}{c|}{38.49} & 37 & \multicolumn{3}{c}{1.132} \\
STL-DivEn & 16,000 & 4,954 & \multicolumn{3}{c|}{35.83} & 35 & \multicolumn{3}{c}{2.424} \\ \hline
            \end{tabular}
        }
        \subcaption{Natural language descriptions statistics: \# unique sentences, \# unique words, \# words per sentences and \# N-gram diversity of natural language descriptions.}
        \label{tab:dataset2}
    \end{minipage}
    
    \vspace{0.3cm} % 增加子表格间距

        \begin{minipage}{\linewidth}
        \centering
        \resizebox{0.95\linewidth}{!}{
            \begin{tabular}{l|cllc|cc|cll}
            \hline
            \multirow{2}{*}{Dataset} & \multicolumn{4}{c|}{\begin{tabular}[c]{@{}c@{}}\#char per \\ identifier\end{tabular}} & \multicolumn{2}{c|}{\begin{tabular}[c]{@{}c@{}}\#digits per\\ constant\end{tabular}} & \multicolumn{3}{c}{\multirow{2}{*}{\begin{tabular}[c]{@{}c@{}}\#identifiers\\ per formula\end{tabular}}} \\ \cline{2-7}
             & \multicolumn{3}{c|}{avg.} & median & \multicolumn{1}{c|}{avg.} & median & \multicolumn{3}{c}{} \\ \hline
            DeepSTL & \multicolumn{3}{c|}{5.50} & 5 & \multicolumn{1}{c|}{2.31} & 2 & \multicolumn{3}{c}{2.59} \\
            STL-DivEn & \multicolumn{3}{c|}{2.63} & 2 & \multicolumn{1}{c|}{1.70} & 2 & \multicolumn{3}{c}{7.2} \\ \hline
            \end{tabular}
        }
        \subcaption{Identifier and constants statistics: \# chars used per identifier, \# number of digits used per constant and \# average number of identifiers per formula.}
        \label{tab:dataset3}
    \end{minipage}
    
    \caption{Dataset statistical analysis of DeepSTL and STL-DivEn.}
    \label{tab:combined_dataset}
\end{table}

% \begin{table}[!t]
%     \centering
%     \resizebox{0.9\linewidth}{!}{
%         \begin{tabular}{lccc}
%             \toprule
%             Dataset & \begin{tabular}[c]{@{}c@{}}Average \\ Subformula \end{tabular} & \begin{tabular}[c]{@{}c@{}}Unique\\ Words\end{tabular} & \begin{tabular}[c]{@{}c@{}}N-gram\\ Diversity Score \end{tabular} \\ \midrule
%             STL-DivEn & 14.66 & 4969 & 2.386 \\
%             DeepSTL & 6.98 & 265 & 1.474 \\ \bottomrule
%         \end{tabular}
%     }
%     \caption{Dataset statistical analysis}
%     \label{tab:dataset}
% \end{table}

Table~\ref{tab:combined_dataset} presents the statistics for the DeepSTL and STL-DivEn datasets.
Specifically, Table~\ref{tab:dataset1} provides statistics on the STL formulas, including subformulas, STL operators, and the N-gram diversity of all STL formulas. A subformula is defined as any well-formed part of a formula that constitutes a complete expression. 
Table~\ref{tab:dataset2} displays statistics for the natural language descriptions, such as the total number of unique sentences, the number of unique words, the average number of words per sentence, and the N-gram diversity of all descriptions. Meanwhile, Table~\ref{tab:dataset3} shows the frequency of identifiers and constants.

The numbers of subformulas and operators in each STL formula indicates that the formulas in the STL-DivEn dataset have more complex structures. 
The total word count of 4,954 unique words in STL-DivEn, compared to only 265 words in DeepSTL, highlights the richer vocabulary in the STL-DivEn dataset. 
Additionally, both the N-gram diversity of the STL formulas and the natural language descriptions demonstrate a greater level of diversity in STL-DivEn.
In conclusion, STL-DivEn is a comprehensive and diverse dataset, making it a valuable resource for further research.

\subsubsection{Ablation Study}

\begin{table}[!t]
    \centering
    \resizebox{0.85\linewidth}{!}{
        \begin{tabular}{lccc}
            \toprule
            Model & \begin{tabular}[c]{@{}c@{}}STL Formula\\ Accuracy\end{tabular} & \begin{tabular}[c]{@{}c@{}}Template\\ Accuracy\end{tabular} & BLEU \\ \midrule
            KGST & 0.5587 & 0.5627 & 0.2142 \\
            - w/o Fine-tuning & 0.5360 & 0.5390 & 0.1978 \\
            - w/o Refinement & 0.4956 & 0.5007 & 0.1784 \\ \bottomrule
        \end{tabular}
    }
    \caption{Ablation experimental results on STL-DivEn.}
    \label{tab:ablationstudy}
    \vspace{-4mm}

\end{table}

To validate the effectiveness of the fine-tuning and refinement modules, we conduct ablation experiments on STL-DivEn, with results shown in Table~\ref{tab:ablationstudy}.
KGST w/o Refinement indicates the KGST model with the Refine module removed, where STL is generated solely by fine-tuning the LLMs.
The results show that when STL is generated using only the fine-tuned LLMs, the metrics are higher than those of the baseline models but lower than those of the complete KGST model.
KGST w/o Fine-tuning indicates the KGST model with the fine-tuning module removed, where STL is generated using only the top five high-similarity NL-STL pairs retrieved from external knowledge as references.
Compared to the complete KGST model, all metrics show a decrease, but still higher than those of the baseline models.
Therefore, we conclude that both fine-tuning and refinement play active roles in STL generation.
% In addition, we also observed that using similarity-based context prompting yields higher scores than using fine-tuning alone, which indicates that the former contributes more to STL generation than the latter.

% 为了验证微调和修正模块的有效性，我们在 STL-DivEn 和 DeepSTL 上进行了消融实验，结果如表所。
% KGST w/o Refine 表示KGST模型移除了Refine模块，仅通过微调大模型的方式生成STL。从结果可以看出，仅使用微调后的大模型生成STL时，各项指标仍高于基线模型，但低于完整的KGST模型。
% KGST w/o Finetune 表示KGST模型移除了微调模块，仅使用从数据库中检索的三个高相似度 NL-STL 对作为上下文提示信息来生成 STL。可以看到，与完整的KGST模型相比，各指标有所下降，但仍高于基线模型。
% 此外，我们还观察到，仅使用微调的分数高于仅使用基于相似度检索的上下文提示的分数，这表明微调对 STL 生成的贡献大于 ICL（基于提示学习）。

\subsubsection{Case Study}

\begin{table}[!t]
    \centering
    \footnotesize
    \resizebox{\linewidth}{!}{
        \begin{tabular}{p{9cm}}  % Set the column width to span the entire table
            \toprule
            \textbf{Case 1:} \\ \midrule
            \textbf{NL (STL-DivEn):} \\
            \begin{minipage}[t]{0.9\textwidth}
                Between time 20 and 50, the sum of signals \(x_1\) and \(x_2\) must not exceed \\ 1.5, unless within 2 to 4 time units later, \(x_3\) exceeds 2.
            \end{minipage} \\ \midrule
            \textbf{GPT4:} \\ 
            \(\G_{[20,50]}(x_1 + x_2 \leq 1.5 \to \F_{[2,4]}(x_3 > 2))\) \\ 
            \textbf{LLaMA 3-8B (Finetuned):} \\
            \(20 \leq t \leq 50 \to ((x_1 + x_2 \leq 1.5) \ \Until{[2,4]} (x_3 > 2)\) \\ 
            \textbf{KGST:} \\
            \(\G_{[20,50]}((x_1[t] + x_2[t] \leq 1.5)  \  \Until{[2,4]} (x_3[t] > 2))\) \\ \midrule
            \textbf{Ground Truth:} \\ 
            \(\G_{[20,50]}((x_1[t] + x_2[t] \leq 1.5)  \ \Until{[2,4]} (x_3[t] > 2))\) \\ \bottomrule
            \\[-0.2cm]
            \toprule

            \multicolumn{1}{l}{\textbf{Case 2:}} \\ \midrule
            \textbf{NL (STL-DivEn):} \\
            \begin{minipage}[t]{0.9\textwidth}
                Whenever signal \(z_2\) falls below -0.5 or exceeds 0.5 within 0 to 500 time \\ units, signal \(z_1\) must exceed 1 within the next 200 time units and maintain \\ that level for at least 50 time units.
            \end{minipage} \\ \midrule
            \textbf{GPT4:} \\ 
            \(\F_{[0,500]}(z_2[t]<-0.5 \vee z_2[t]>0.5) \to (\F_{[0,200]}(z_1 > 1) \land \G_{[0,50]}(z_1 > 1))\) \\ 
            \textbf{LLaMA 3-8B (Finetuned):} \\
            \(\F_{[0,500]}((z_2[t] < - 0.5 \vee z_2[t] > 0.5) \to \F_{[0,200]}\G_{[0,50]}(z_1 > 1))\) \\ 
            \textbf{KGST:} \\
            \(\G_{[0,500]}((z_2[t]<-0.5 \vee z_2[t]>0.5) \to \F_{[0,200]}\G_{[0,50]}(z_1[t]>1))\) \\ \midrule
            \textbf{Ground Truth:} \\ 
            \(\G_{[0,500]}((|z_2| > 0.5) \to (\F_{[0,200]}\G_{[0,50]}(z_1 > 1)))\) \\ \bottomrule
        \end{tabular}
    }
    \caption{Generated STL formulas from different models on STL-DivEn.}
    \label{tab:casestudy}
    \vspace{-5mm}
\end{table}

To intuitively demonstrate how KGST improves the quality of STL generation, we present a case study in Table~\ref{tab:casestudy}.
In this study, we compare the STL formulas generated by KGST with those generated by GPT-4 and the fine-tuned LLaMA 3-8B model.
% and compare the STL formulas generated by KGST with those generated by GPT-4 and the fine-tuned LLaMA 3-8B model to intuitively demonstrate the improvement of KGST in enhancing the quality of STL generation.
In Case 1, according to the natural language description, $x_3 > 2$ must occur within $2$ to $4$ time units in the future. However, GPT-4 incorrectly uses $\F_{[2,4]}(x_3 > 2)$ to express a logical "until", which is not accurate. 
On the other hand, while the syntax of LLaMA 3-8B is not fully compliant (e.g., it does not explicitly use $\G_{[20,50]}$ to indicate the global time interval constraint), its basic logic is correct.
In Case 2, both GPT-4 and LLaMA 3-8B use incorrect syntax for the triggering condition. 
The correct expression should use the global operator $\G_{[0,500]}$ to specify that the triggering condition must be monitored across the entire time interval, rather than at a specific point in time. 
Furthermore, in the formula generated by GPT-4, $\F_{[0,200]}(z_1 > 1)$ and $\G_{[0,50]}(z_1 > 1)$ are used in parallel, but there is no indication of the sequential relationship. 
The correct logic should specify that $z_1 > 1$ must first occur, followed by its persistence for $50$ time units.
These results confirm that KGST effectively corrects errors in the generated STL, such as misused operators or invalid syntax.

% 我们在表X中展示了一个研究案例，并与GPT4生成的和微调后的Llama3生成的进行对比，以更直观地说明KGST在提升Signal Temporal Logic生成质量方面的改进。
% 在第一个案例中，根据自然语言，要求的是在未来2到4时间单位内，x_3>2必须发生，而GPT-4错误使用F[2,4](x_3>2)表达“直到”的逻辑。而LLaMA 3-8B的语法不够规范，没有明确用G[20,50]表达全局时间区间的约束，但逻辑是正确的。
% 在第二个案例中。GPT-4和Llama3的触发条件语法均使用错误，应该使用全局变量G[0,500]，明确在整个时间区间内检测触发条件，而非仅检测某一时刻是否触发。此外，在GPT-4生成的公式中，将F[0,200](z_1>1)和G[0,50](z_1>1)写作并列关系，缺乏递进性，即先到达z_1>1，然后持续50时间单位。
% 这些结果验证了KGST能够有效地修正生成的STL中存在的语法或语义上的错误，如纠正时序运算符错误，或改正错误的语法。

\subsubsection{Impact of Refinement}

\begin{figure}[!t]
    \centering
    \includegraphics[scale=0.35]{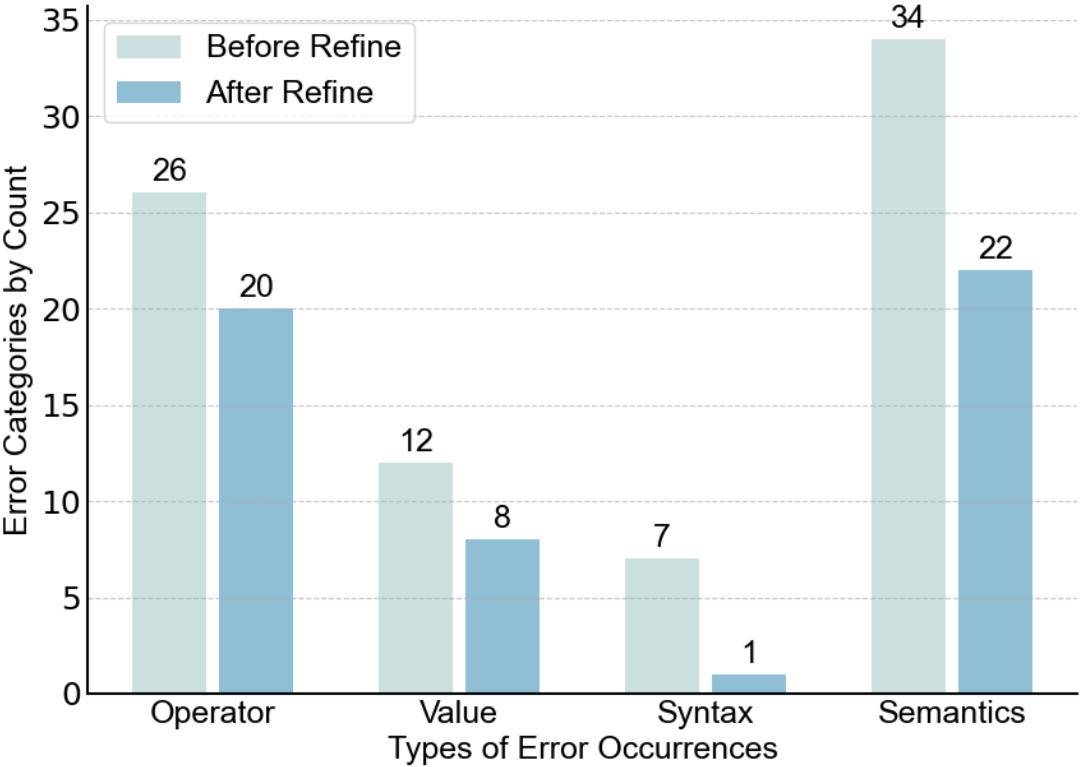}
    
    \caption{Tracking errors before and after refinement.}
    \vspace{0mm}
    \label{fig:refinement_difference}
\end{figure}

To validate the impact of refinement on specific error types, we track four types of errors in 100 generated STL formulas: incorrect operator usage, value errors, syntax violations, and semantic inconsistencies with the corresponding NL. The differences before and after the refinement process are shown in Figure~\ref{fig:refinement_difference}, and it is observed that the frequencies of all error types have decreased.

\begin{figure}[!t]
    \centering
    \includegraphics[scale=0.31]{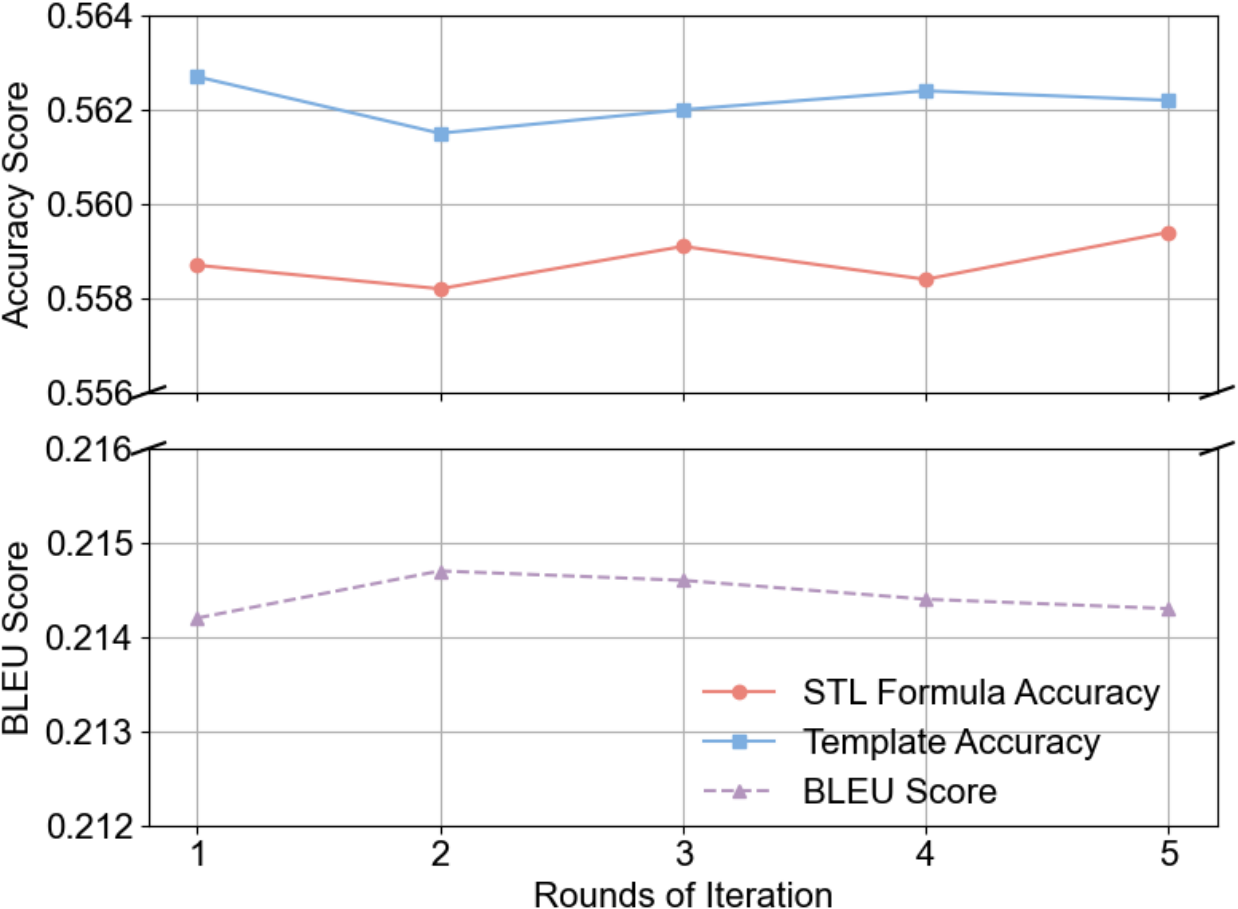}
    \caption{Impact of iteration rounds on refinement.}
    \vspace{-2mm}
    \label{fig:refinement_iteration}
\end{figure}

We also conduct an experimental analysis of the iteration rounds by calculating the STL Formula Accuracy, Template Accuracy, and Bleu score for different numbers of refinement iterations on STL-DivEn. 
Figure~\ref{fig:refinement_iteration} shows that as the number of iterations increases, there is no significant impact on the effect of refinement, because each iteration uses the same NL-STL as the reference.

    % Subfigure 2
% \caption{Analysis of refinement effects.}
% \label{fig:refinement_comparison}

% \begin{figure}[!t]
%     \centering
%     % Subfigure 1
%     \begin{subfigure}[b]{1\linewidth}
%         \centering
%         \includegraphics[scale=0.35]{Pictures/refinement_difference.pdf}
        
%         \caption{Tracking errors before and after refinement.}
%         \vspace{1mm}
%         \label{fig:refinement_difference}
%     \end{subfigure}
%      %\vspace{1mm} % Space between the two subfigures

%     % Subfigure 2
%     \begin{subfigure}[b]{1\linewidth}
%         \centering
%         \includegraphics[scale=0.31]{Pictures/refinement_iteration.pdf}
%         \caption{Impact of iteration rounds on refinement performance.}
%         \vspace{-2mm}
%         \label{fig:refinement_iteration}
%     \end{subfigure}
%     \caption{Analysis of refinement effects.}
%     \label{fig:refinement_comparison}
%     %\vspace{-1mm}
% \end{figure}

% 我们还统计了错误的类型：运算符使用错误，数值错误，语法不符合规则，语义与对应的NL不一致。在Refine前后的区别。如表所示。从中可以看出，对于语法不符合规则一项做的最好。其余的几类错误也都得到了改善。

% 我们通过计算不同的Refine次数在STL-DivEn和DeepSTL上的STL Formula Accuracy，Template Accuracy和Bleu score，进行迭代次数的试验分析。从图中可以看出，随着迭代轮次的增加，对Refine的效果没有显著的影响，因为每次迭代都是用相同的NL-STL作为范式进行Refine。

\subsubsection{Scaling Effect}

\begin{figure}[!t]
    \centering
    \includegraphics[width=1\linewidth]{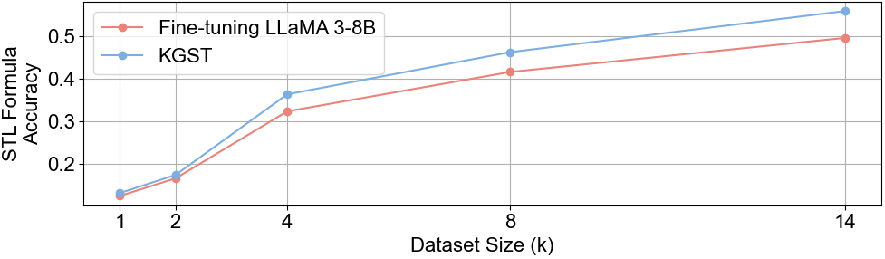}
    \vspace{-5mm}
    \caption{Scaling effect of STL-DivEn dataset on STL formula accuracy.}
    \vspace{-5mm}
    \label{fig:scaling_law}
\end{figure}

Figure~\ref{fig:scaling_law} presents the results of the scaling effect experiments on the STL-DivEn dataset. 
It illustrates how STL Formula Accuracy changes as the dataset size increases. 
Both the fine-tuning and KGST show gradual improvement with the growth of the dataset, with KGST consistently outperforming the fine-tuning across all dataset sizes, particularly on larger datasets.
The performance on other evaluation metrics can be found in Appendix~\ref{sec:sca_law_full}.
% 我们在STL-DivEn数据集上进行了Scaling Law的实验，实验结果如图。图展示了STL Formula Accuracy、Template Accuracy 和 BLEU Score 随数据集规模扩展的变化趋势，Fine-tune和KGST的表现都随着数据集规模的增加而逐步提高，并且，KGST所有数据集规模下都优于 Fine-tune，尤其是在大规模数据上表现出明显的优势。

%% file: 6Conclusion.tex
\section{Conclusion}

%In this work, we present a new dataset, STL-DivEn, which features NL-STL pairs with enhanced diversity. Each pair has been manually verified for accuracy. Based on this dataset and DeepSTL, we conduct transformation experiments. Specifically, we perform Supervised Fine-Tuning (SFT) on Llama-3-8B, using instruction tuning. Additionally, we introduce a novel refinement phase, culminating in the development of the KGST model, designed to optimize STL transformation during the initial stage. Results from both metric-based evaluations and human evaluations demonstrate that our approach significantly improves transformation capabilities across two datasets. Our approach facilitates the automatic extraction of temporal and continuous constraints in AI-embedded systems, supporting efficient and reliable modeling to ensure the safety and robustness of these systems.

% 在这项工作中，我们提出了一个名为 STL-DivEn 的新数据集，该数据集包含具有多样性增强的自然语言-STL 对。每个对都经过了人工验证以确保准确性。在此数据集和 DeepSTL 的基础上，我们进行了翻译实验。具体而言，我们在 Llama-3-8B 模型上进行监督微调（SFT），并采用指令调优。此外，我们引入了一个新的优化阶段，最终开发了 KGSR 模型，用于优化第一阶段的 STL 合成。基于指标的评估和人工评估的结果表明，我们的方法在两个数据集上的翻译能力显著提升。

In this work, we present a new dataset, STL-DivEn, which features NL-STL pairs with enhanced diversity. Additionally, we introduce the KGST framework, a novel approach for transforming natural languages into STL. Results from both metric-based evaluations and human evaluations demonstrate that our approach significantly improves transformation capabilities across two datasets. Our approach facilitates the automatic extraction of temporal and continuous constraints in cyber-physical systems, supporting efficient and reliable modeling to ensure the safety and robustness. 
% of these systems.

%% file: Appendix.tex
% \onecolumn

\section{Prompts input to Large Language Models}
In this section, we present the prompts designed to guide large language models.

\subsection{Prompts for GPT-4 to generate NL-STL pairs}

\begin{figure}[!htb]
    \centering
    \includegraphics[width=1\linewidth]{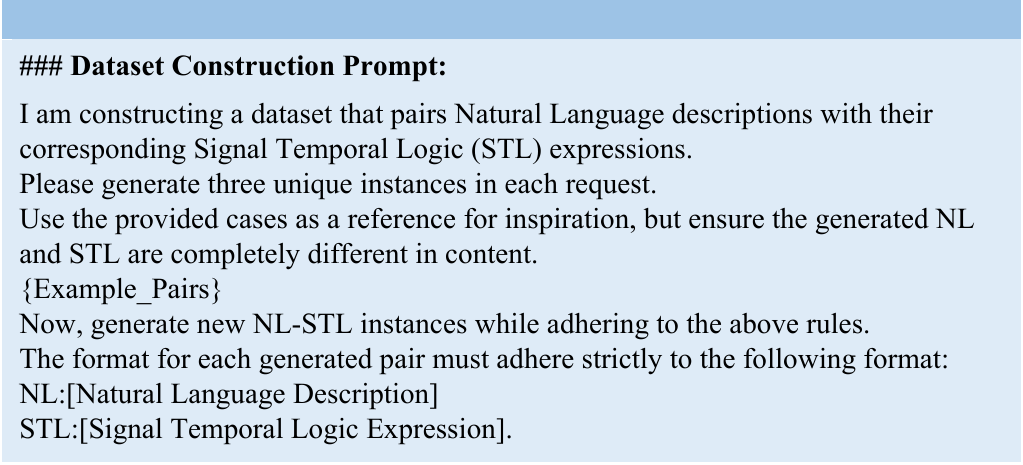}
    %\vspace{-1mm}
    \caption{Evolution Prompts for GPT-4 in NL-STL Pairs  Generation.
    }
    \label{fig:Datast_Prompt}
    \vspace{-5mm}
\end{figure}

Figure~\ref{fig:Datast_Prompt} shows the prompt used for GPT-4 to generate NL-STL pairs. The example pairs are selected from the seed set using a clustering algorithm.

\subsection{Prompts for LLMs to Generate STL}
\label{sec:GPT-4_generate}
\begin{figure}[!htb]
    \centering
    \includegraphics[width=1\linewidth]{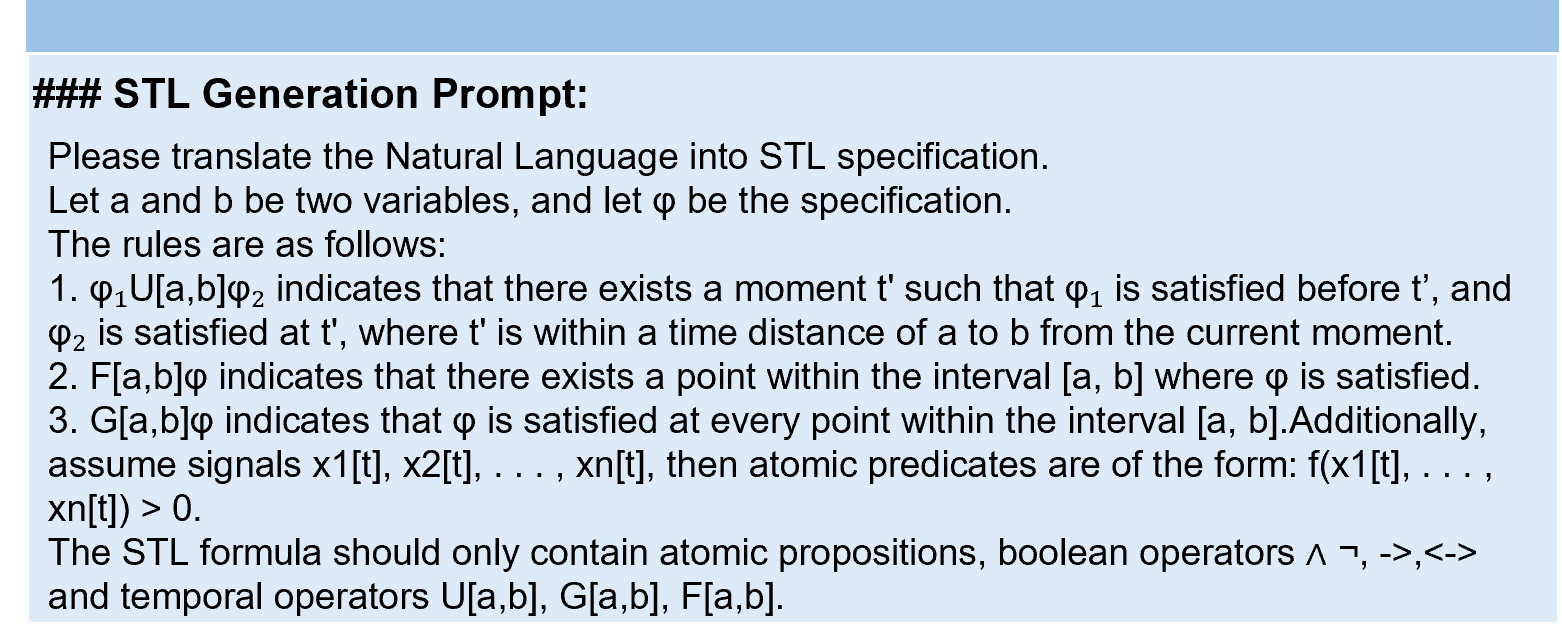}
    %\vspace{-1mm}
    \caption{The prompt for Baseline Models to generate STL formulas.
    }
    \label{fig:STLGeneration}
    \vspace{-5mm}
\end{figure}

Figure~\ref{fig:STLGeneration} shows the prompts used for baseline models, including GPT-3.5, GPT-4, and DeepSeek, to generate STL formulas from input natural language descriptions.

%\newpage
\subsection{Prompts for Refinement Part in KGST}
\label{sec:KGST_refine}
\begin{figure}[!htb]
    \centering
    \includegraphics[width=1\linewidth]{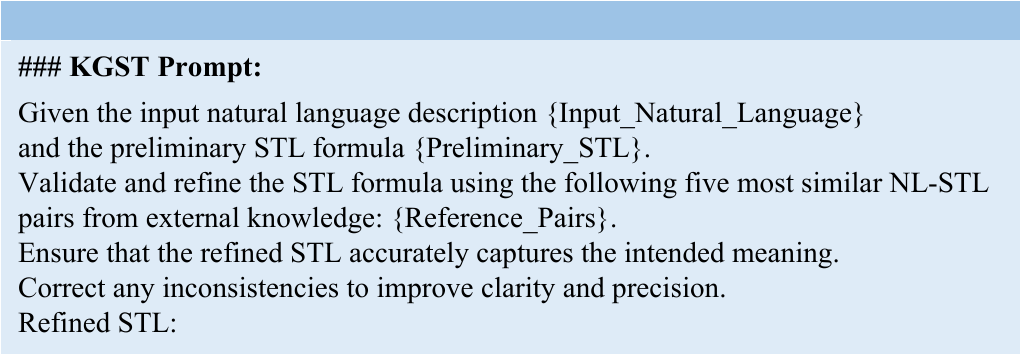}
    %\vspace{-1mm}
    \caption{The prompts in the refinement part of KGST.
    }
    \label{fig:KGST_Refine}
    \vspace{-5mm}
\end{figure}

Figure~\ref{fig:KGST_Refine} shows the prompts used for KGST to refine the preliminary STL. Reference pairs refer to the top $K$ NL-STL pairs selected from external knowledge based on their similarity to the transformed natural language.

\begin{figure}[!htb]
    \centering
    \includegraphics[width=1\linewidth]{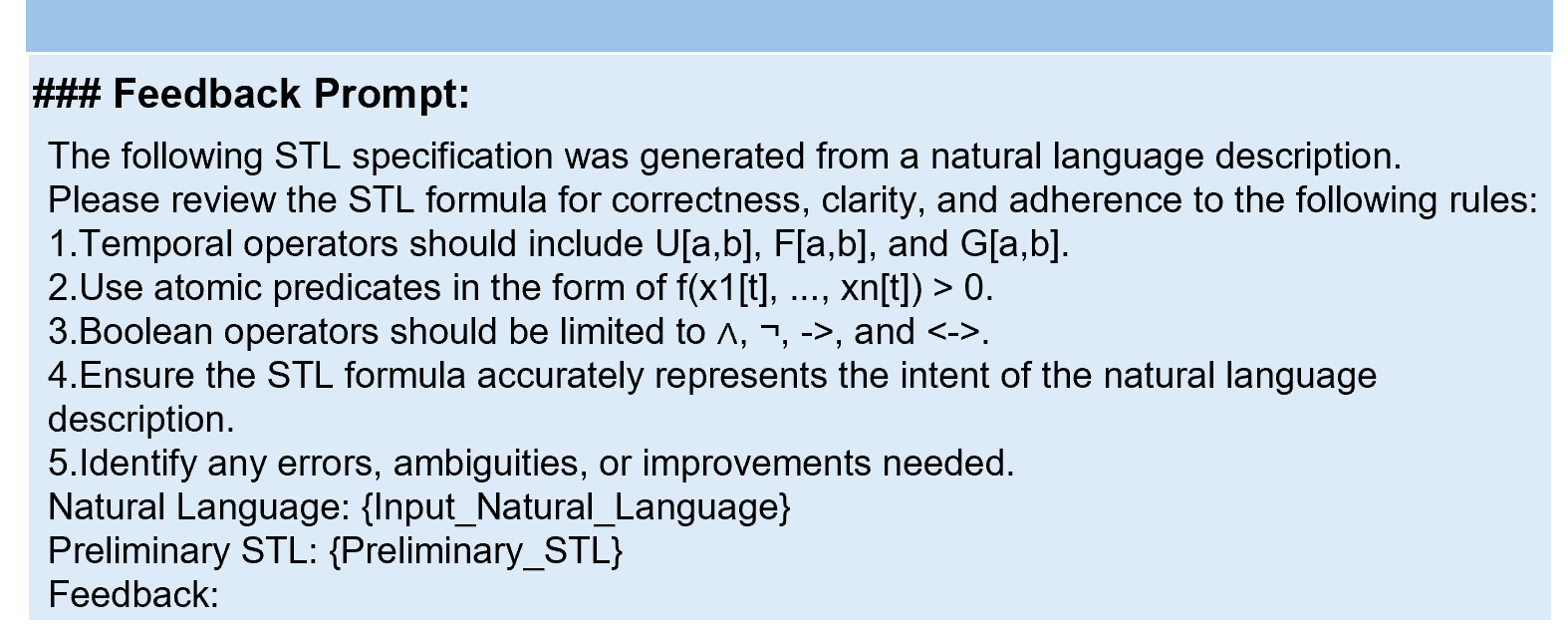}
    %\vspace{-1mm}
    \caption{The prompts in the feedback part of Self-Refine.
    }
    \label{fig:SelfRefine_feedback}
    \vspace{-5mm}
\end{figure}

\subsection{Prompts for Self-Refine}
\begin{figure}[!htb]
    \centering
    \includegraphics[width=1\linewidth]{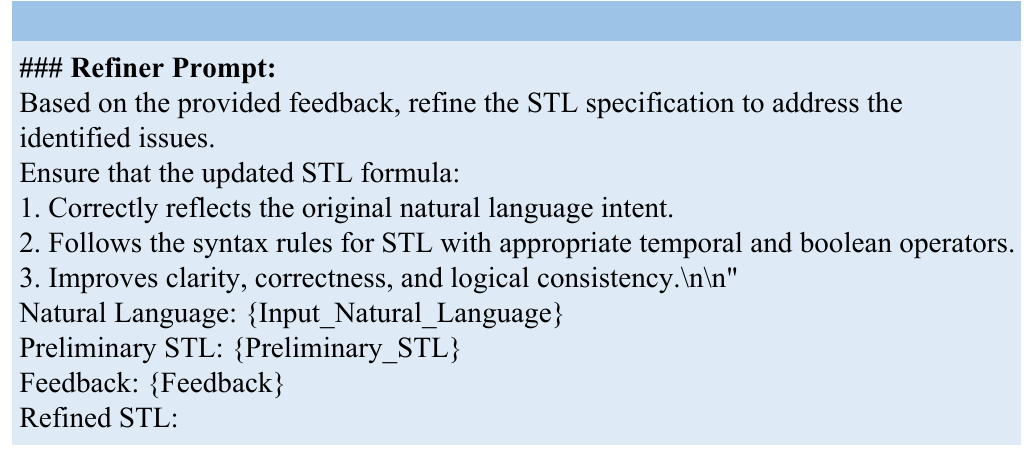}
    %\vspace{-1mm}
    \caption{The prompts in the refinement part of Self-Refine.
    }
    \label{fig:SelfRefiner}
    \vspace{-5mm}
\end{figure}

Figure~\ref{fig:SelfRefine_feedback} shows the prompts used for GPT-4 to generate feedback on whether the STL is correct based on the STL generation criteria for the given natural language input and its corresponding STL. 
Figure~\ref{fig:SelfRefiner} shows the prompts used for GPT-4 to refine the preliminary STL based on the feedback.

%\newpage
\subsection{Prompts for KGST w/o Finetune}
\begin{figure}[!htb]
    \centering
    \includegraphics[width=1\linewidth]{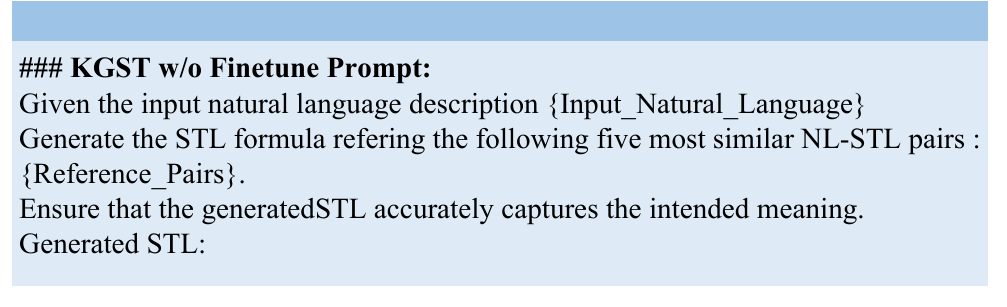}
    %\vspace{-1mm}
    \caption{The prompts for KGST w/o Finetune to generate STL.
    }
    \label{fig:InContextPrompt}
    \vspace{-5mm}
\end{figure}

Figure~\ref{fig:InContextPrompt} shows the prompts used for GPT-4 to generate STL based on the input natural language description and the top $K$ NL-STL pairs retrieved from external knowledge with the highest similarity to the input, which serve as reference pairs in the context.

% \newpage
\section{Evaluation Metrics}
\label{sec:Metrics}
STL formula accuracy ($A_F$) and template accuracy ($A_T$). The first metric measures the alignment accuracy between the reference and predicted sequences at the string level, while the second metric involves transforming both the reference and predicted instances into STL templates and then calculating their alignment accuracy. For example: 

Formula: $ \text{eventually} \, (a < 5) \Rightarrow \, \text{Template}: \, \text{G} \, (\phi) $    

Formula: $ \text{eventually} \, (b < 5) \Rightarrow \, \text{Template}: \, \text{G} \, (\phi) $

The first line represents the reference sequence, and the second line corresponds to the model's prediction. 
To illustrate more clearly, spaces are inserted between each token, resulting in six tokens in the formula and four tokens in the template.
In the formula, five tokens appear in the same positions—‘$G$’, ‘$($’, ‘$<$’, ‘$5$’, ‘$)$’—while the remaining token ‘$a$’ in the reference is mistranslated as ‘$b$’. Therefore, the formula accuracy ($A_F$) is calculated as:

$$
A_F = \frac{5}{6}
$$

For the template, since all tokens align perfectly, the template accuracy ($A_T$) equals:

$$
A_T = 1
$$
% \newpage
\section{Details of Implementation}
\label{sec:Details}
The experiments are conducted on eight NVIDIA 4090 GPUs, with all implementations utilizing PyTorch\footnote{https://pytorch.org/}, LLaMA-Factory\footnote{https://github.com/hiyouga/LLaMA-Factory}, and Huggingface's Transformers\footnote{https://github.com/huggingface/transformers}. To ensure efficient training, the learning rate is set to 5e-5 and the batch size is 16. To ensure the adequacy of the training results, the model is run for 10 epochs under each setting.

\newpage
\section{Scaling Effect of Multi-Metrics}
\label{sec:sca_law_full}

\begin{figure}[!t]
    \centering
    \includegraphics[scale=0.5]{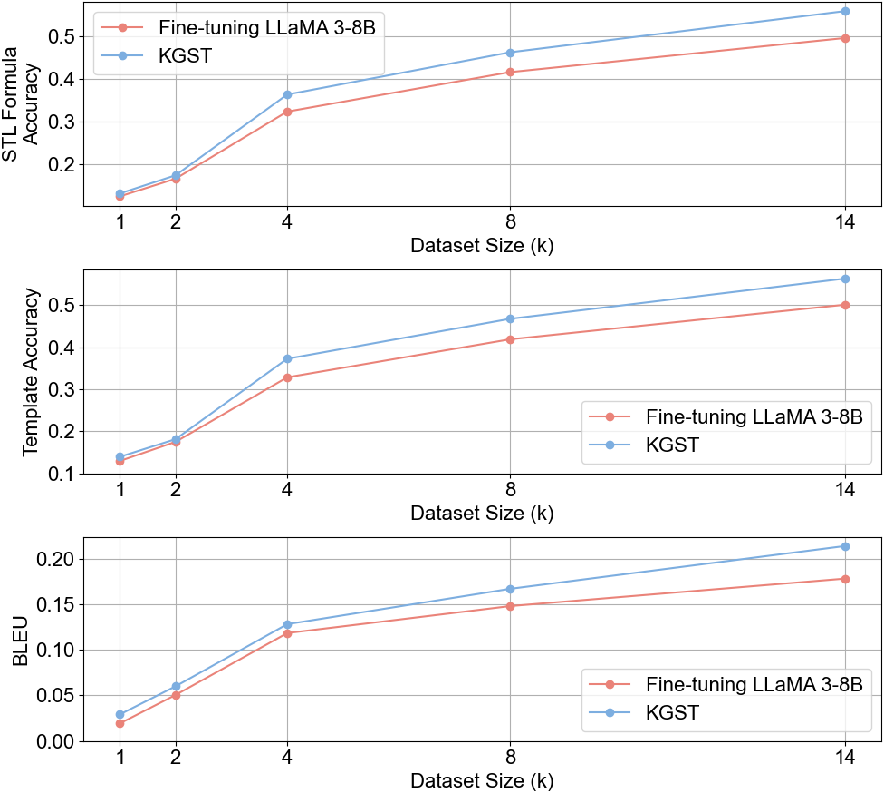}
    %\vspace{-1mm}
    \caption{Scaling effect of STL-DivEn on three evaluation metrics.
    }
    \label{fig:scaling_law_ful}
    %\vspace{-5mm}
\end{figure}

Figure~\ref{fig:scaling_law_ful} shows the scaling effect of the STL-DivEn dataset, illustrating the performance metrics of STL generation after fine-tuning with Llama-3-8B on the STL-DivEn dataset, as well as the performance of KGST in generating STL formulas. The metrics include STL formula accuracy, template accuracy, and BLEU score, as the dataset size increases from 1k to 16k.